\documentclass[11pt]{article}
\usepackage{geometry}
\usepackage{coling2020}
\usepackage{times}
\usepackage{url}
\usepackage{latexsym}
\usepackage{microtype}
\usepackage{algorithm}
\usepackage{algorithmic}
\usepackage{graphicx}
\usepackage{color}

\colingfinalcopy 

\title{CS-Embed at SemEval-2020 Task 9: The effectiveness of code-switched word embeddings for sentiment analysis}

\author{Frances A. Laureano De Leon \\
  University of Birmingham \\
  United Kingdom \\
  {\tt \scriptsize{laureanofa@gmail.com}} \\
  \And
  Florimond Gu{\'e}niat \\
  Birmingham City University \\
  United Kingdom \\
  {\tt \scriptsize florimond.gueniat@bcu.ac.uk} \\
  \And
Harish Tayyar Madabushi \\
  University of Birmingham \\
  United Kingdom \\
  {\tt \scriptsize H.TayyarMadabushi.1@bham.ac.uk} \\ }

\date{}

\begin{document}
\maketitle
\begin{abstract}
   The growing popularity and applications of sentiment analysis of social media posts has naturally led to sentiment analysis of posts written in multiple languages, a practice known as code-switching. While recent research into code-switched posts has focused on the use of multilingual word embeddings, these embeddings were not trained on code-switched data. In this work, we present word-embeddings trained on code-switched tweets, specifically those that make use of Spanish and English, known as Spanglish. We explore the embedding space to discover how they capture the meanings of words in both languages.  We test the effectiveness of these embeddings by participating in SemEval 2020 Task 9: ~\emph{Sentiment Analysis on Code-Mixed Social Media Text}. We utilised them to train a sentiment classifier that achieves an F-1 score of 0.722. This is higher than the baseline for the competition of 0.656, with our team (codalab username \emph{francesita}) ranking 14 out of 29 participating teams, beating the baseline. 
  

\end{abstract}

\section{Introduction}
\label{intro}

%
%
\blfootnote{
    %
    %
    %
    %
    \hspace{-0.65cm}  
    This work is licensed under a Creative Commons 
    Attribution 4.0 International Licence.
    Licence details:
    \url{http://creativecommons.org/licenses/by/4.0/}
    %
    %
}

Sentiment Analysis (SA), or opinion mining, 
aims to identify subjective information from text. 
SA plays a pivotal role in promoting and identifying reception of products in platforms such as Amazon, or identifying the sentiment and opinions on numerous topics on social media platforms, such as Twitter or Facebook.
As a consequence, SA is a popular task in the Natural Language Processing (NLP) community.

Sentiment analysis of social media posts has been a popular research topic, and has naturally led to the sentiment analysis of posts in languages other than English, as well as code-switched posts. 
Written text in which multiple languages co-exist, known as code-switching or code-mixing, is now more abundant thanks to the ample use of social media.
Recent research on code-switched text has focused on using multilingual word embeddings, however, these embeddings were not trained using code-switched text. In this work, we present word-embeddings trained on code-switched social media text, which serves as our main contribution for SemEval 2020 Task 9~\footnote{https://competitions.codalab.org/competitions/20789}.

This task consisted of classifying tweets into one of three classes: positive, neutral, or negative. 
The dataset used for this task is the one provided by the competition organisers, who collected and annotated the corpus,~\cite{patwa2020sentimix}. 
We trained a BiLSTM~\cite{Schuster} classifier to detect the sentiment of code-mixed tweets by creating our own code-switched embeddings. Using the techniques outlined in section~\ref{system_overview}, large amounts of code-switched text can be collected to train code-switched embeddings for different NLP applications, such as semantic processing, dependency parsing~\cite{DBLP:journals/corr/abs-1902-09492}, as well as sentiment analysis, all of which have been identified as challenges posed by code-mixing in NLP tasks~\cite{patwa2020sentimix}. We release code for the code-switched sentiment classifier, crawled Twitter data, and associated experimental data, including hyper-parameters~\footnote{~\url{https://github.com/francesita/CS-Embed-SemEval2020}}. 

\section{Background}

To our knowledge, there is a lack of literature and work on code-switched (CS) word embeddings.
There is one paper on word embeddings using synthetic code-switched data~\cite{Sitaram2018}. 
There is, however, considerably more literature on bilingual word embeddings created on parallel monolingual corpora. These are often used in cross-lingual and bilingual tasks~\cite{Huang2015,Zhang2018}.

Numerous methods have been introduced for creating bilingual word-embeddings.
For instance, canonical correlation analysis (CCA) finds the association or correlation between two vectors,~\cite{Livescu2015} to create bilingual word embeddings.
Bilingual Canonical Correlation Analysis (BiCCA) was also developed, it is a technique in which monolingual embeddings are mapped to the same space by use of CCA,~\cite{Faruqui2014}.
Not long after, BiCCA was extended to the learning of non-linear deep canonical correlation (DCCA), which often outperforms linear CCA~\cite{Livescu2015}.

This method was followed by the introduction of Bilingual Skip-gram, in which bilingual representations of words are learned from scratch,~\cite{Manning2015}.
~\newcite{Huang2015} proposes learning bilingual sentiment word embeddings (BSWE) for English-Chinese SA by use of labelled documents and their translation rather than using parallel corpora with results that outperformed state of the art at the time.

\newcite{Agirre2017} proposed to train embeddings individually, in their own monolingual space, and to map one monolingual embedding to the space of the other by use of a linear transformation.
This technique makes use of bilingual dictionaries with as little as 25 words.

However, recent progress shows that bilingual methods of creating word embeddings are not well-suited for code-mixed tasks,~\cite{Sitaram2018}.
Indeed, there are grammatical structures that are not captured by monolingual text that exist when code-mixing occurs~\cite{Lipski2005,Sankoff1981}. 
It appears that bilingual embeddings created from monolingual or synthetic data may not be well suited for studies of code-mixed text.
By training CS word embeddings, words that are usually used together in the same space when language switching occurs, will be clustered together~\cite{D.Jurafsky2019}.
We believe code-switched embeddings will be more robust in classifying CS text than others because the patterns in language when code-switching occurs will be captured in the embeddings. 
For these reasons, we have decided to train word embeddings on code-switched data.

%
%

\section{System Overview and Experimental Set-up}
\label{system_overview}

We chose to train our own multilingual word embeddings using code-switched social media text because we hypothesise that these embeddings will be more effective in downstream tasks than combining monolingual embeddings from two different languages. To this end, we collected code-switched tweets in Spanglish~\footnote{~\url{https://github.com/francesita/CS-Embed-SemEval2020}}. 
The external tools utilised for this work are Gensim's word2vec model~\footnote{\url{https://radimrehurek.com/gensim/models/word2vec.html}}, natural language toolkit (NLTK)~\footnote{\url{https://www.nltk.org/}}, keras version 2.2.4~\footnote{\url{https://pypi.org/project/Keras/}}, and tensorflow version 2.0~\footnote{\url{https://www.tensorflow.org/}}. 

\subsection{Twitter Data Collection}
\label{twiiter_collection}
In order to train the code-switched word embedding\footnote{see~\url{https://github.com/francesita/Code-switch-Embeddings-for-Sentiment-Analysis}}, over 1,000,000 tweets were collected using an in-house code  between 12 July, 2019 and 31 August, 2019 and between December 2019 and February 2020.
Tweepy was used to gather the tweets with the Cursor function which allows for tweets to be searched for by query word, language, location and date.

First, a text document containing 315 words in Spanish and Spanglish was created and used as a code-switch key word list.
The Spanish words chosen are part of a list of 500 most commonly used Spanish words according to the Dictionary of the Royal Spanish Academy. 
This list of 500 words was cross-referenced with a Portuguese dictionary in order to remove any words in common, as both these languages are romance languages.
Words containing less than four letters were removed, to decrease the chances of overlap with other languages sharing words with Spanish, such as Tagalog. 
Proper nouns were removed as well.
The Spanglish words included in the list refer to words that are hybrids of Spanish and English, such as the word \emph{janguear} which both means and stems from the English phrase \emph{hang-out} and ends with~\emph{-ar}, which allows the word \emph{janguear} to be conjugated as a typical~\emph{-ar} infinitive Spanish verb.
We looked-up the most commonly used Spanglish words in order to include these in the key word list.
Noteworthy, many of the Spanglish key words included are typically used in Puerto Rico and Florida, which may induce bias, though efforts were made to include Spanglish words used by Hispanics living in Texas, California and other parts of the United States. 

The words in the code-switched document were used as the query word in the cursor function in tweepy, while the language of the tweets extracted was English. 
Resulting collected tweets were considered code-mixed and were used to train the word-embeddings. 

\subsection{Preprocessing}
NLTK was used for preprocessing the collected tweets.
The tweets were tokenized. 
Spanish and English punctuation, as well as stop words were removed. 
We also extended contractions in English (such as can't to cannot). 
Emojis were kept, as these are useful for training sentiment classifiers~\cite{Lo2017}.

\subsection{Creation of Word-embeddings}
\label{WEcreation}

Twitter data collected using methods described in~\ref{twiiter_collection} were used to train code-switched word-embeddings using Gensim's word2vec model,~\cite{Mikolov2013}. 
We decided to train our embeddings using the CBoW algorithm rather than skip gram as it is better suited to smaller datasets,~\cite{Zhang2018}. 
The embeddings were trained for 20 epochs with size 100, window of 5,  and 10 workers for 30 iterations.

\subsection{Description of BiLSTM Classifier}
A BiLSTM model was used to classify the sentiment of the tweets.
It was trained on 12,002 tweets, validated on 2,998 tweets and tested on 3,789 tweets. 
The BiLSTM model contained one embedding layer, in which the word embeddings we created were utilised. 
It had three bidirectional layers with dropout of 0.2, two dense layers, one with dimension 100 and dropout 0.3 and the other with dimension 3. 
The activation function used was \emph{relu} for all layers except for the output classification layer, for which the \emph{softmax} activation function was used.
Adamax was used as the optimiser for the model, with learning rate 0.0002.
Early Stopping was also used with min\_delta 0.0002 and patience 5.

\section{Results}
\label{results}
As previously mentioned, our team decided to train word embeddings with code-switched text as we believed that these embeddings would be more effective than other multilingual embeddings for downstream tasks.
Our team results and ranking in Task 9 are presented in table~\ref{task_results}. 

\begin{table*}\caption{SelEval2020 Task 9 Results}
\label{task_results}
\center
\begin{tabular}{|l||ll|}
\hline
Rank        & Users                           & Best Score     \\
\hline
1           & LiangZhao                       & 0.806          \\
2           & rachel                          & 0.776          \\
3           & asking28                        & 0.756          \\
            &                                 &                \\
\textbf{14} & \textbf{francesita (this work)} & \textbf{0.722} \\
            & ...                             &                \\
23          & suraj1ly (\textbf{organiser baseline})   & 0.656          \\
\hline
\end{tabular}
\end{table*}

\subsection{Embedding Evaluation}
\label{evaluation}

In the initial stages of data gathering to make code-switched word embeddings, the ten most similar words to each of the key words were used to do some simple evaluations on the embeddings.
This allowed us to identify some issues with early data collection methods, such as the inadvertent collection of code-switched tweets in Portuguese, Italian, and Tagalog mixed with English.
Our embeddings were also biased to current events at the time of data collection.
Table~\ref{tab:prob_embed} shows examples of what issues were present in early embeddings. 
The letter in between brackets on the table indicate a language, or  a pop culture icon (denoted by \emph{I}). 

\begin{table*}\caption{Problematic Word Mappings}\label{tab:prob_embed}
\centering
\resizebox{\textwidth}{!}{\begin{tabular}{|l||l|l|}
    \hline
    Word &  Most Similar Words  & Problem\\
    \hline
    calor    & anitta(\emph{I}), muitocalor(\emph{port}), ozuna(\emph{I}), muito(\emph{port}), gosto(\emph{port}) &       language and current events\\
    gracias  &  thanks(\emph{en}), obrigada(\emph{port}), muchas(\emph{es}), grazie(\emph{it}), bendiciones(\emph{es})   &       language \\
    amazon   &  rainforest(\emph{en}), fires(\emph{en}), deforestation(\emph{en}), wildfires(\emph{en}), brazil(\emph{en}) & current events\\ 
    \hline
\end{tabular}}
\end{table*}

We believe early issues were caused by the limit of tweet extractions set forth by Twitter (there is a 7 day limit the amount of data that can be collected at one time) as well as the limited number of key words used in early stages.
The embeddings, after all, are only as good as the amount of data used to train them.

Happenings and current events of the time also influenced the embedding space to some degree as seen in table~\ref{tab:gobe}.
Many of the words surrounding \emph{gobernador} are related almost exclusively to Puerto Rico's government, which was having protests during the time of data collection.
This shows that the time-frame in which data was collected allowed for bias to exist in the embeddings.

\begin{table}[h]\caption{Most Similar Words to \emph{gobernador} (Governor)}
\label{tab:gobe}
\center
\begin{tabular}{|l||l|}
\hline
Gobernador (governor) & \begin{tabular}[c]{@{}l@{}}renuncie (resign), gobernadora (female governor), renunciar (to resign), \\abajorosello (down with Rosello), cabron (expletive bastard/ cuckold)\end{tabular} \\ \hline
\end{tabular}
\end{table}

Word embeddings may be biased in terms of what words are clustered together, which can be seen with the word \emph{presidente} in table~\ref{biling_rep}.
It is evident that most of the words, and proper nouns clustered around the word \emph{presidente} are associated to the political systems and leaders in the Americas.

\begin{table*}[h!]\caption{Examples of Bilingual Mapping from CS Embeddings}
\label{biling_rep}
\center
\begin{tabular}{|l||l|}
\hline
Word                   & Similar Words                                                                  \\
\hline
Abuelo (grandfather)    & abuela (grandmother), grandpa, abuelita (grandma),\\
                        & dad, grandma, cousin, grandparents    \\
Novio (boyfriend)       & esposo (husband), boyfriend, novia (girfriend), \\
                        & aver (to have), bf (boyfriend), gf (girlfriend) \\
Presidente (President)  & president, dictator, presidentethe, presidentes, pres, guaido, reelection      \\
\hline
\end{tabular}
\end{table*}

We believe that as most of the code-switching between Spanish and English occurs in the Americas, the word embeddings reflect this.
As we continue to gather data, much of the bias should disappear, as it is present due to the time-frame in which tweets were gathered as shown by table~\ref{early_late_embed}.

\begin{table}[h!]\caption{Early and Later embeddings for the word \emph{Calor}}
\label{early_late_embed}
\begin{tabular}{|l||l|}
\hline
Calor (Hot)      & Most Similar Words                                        \\
\hline
Early Embedding  & anitta (artist), muitocalor (Portuguese), ozuna (artist), muito (Portuguese) \\
Latest Embedding & frío (cold), piscina (pool), sed (thirst), agua (water)                      \\
\hline
\end{tabular}
\end{table}

However, it is likely that not all bias will disappear, as code-switching occurs regionally, and data collected from these regions will likely show bias to that part of the world.
For example, if code-switched data is collected from India, the concerns of the people, political or otherwise, are likely to be very different from the concerns of someone living in the Americas. 
Since code-switched word embeddings will be used for specific languages where code-switching is prominent, the embeddings need only reflect the happenings of that region.

\subsection{Quantitative Analysis}
 We created a confusion matrix to understand where the system misclassified the sentiment of tweets. 
 We utilised the development dataset as the labels for the test data are not yet released. 
 As can be seen in table~\ref{conf_matrix}, neutral tweets were the most challenging to classify.
 These tweets were overwhelmingly classified as positive by our model. In the future, we plan on comparing the output of the classifier, and to use concurrent models to prevent miscalssification of neutral tweets into the positive class.
 It can also be seen that the negative class is well identified in our model, with no misclassified negative tweets, see table~\ref{conf_matrix}, and that the positive class is classified correctly for the most part. 
 Results can be found Table~\ref{compare_results}.
 
\begin{table*}[h!]\caption{Confusion Matrix. Rows are actual classes while columns are predicted classes.}
\label{conf_matrix}
\center
\begin{tabular}{|l||ll|}
\hline
         & Not Positive & Positive \\
\hline
True Not Positive & 860          & 639      \\
True Positive     & 457          & 1042    \\
\hline
        & Not Neutral & Neutral \\
\hline
True Not Neutral & 1548        & 457     \\
True Neutral     & 639         & 354      \\
\hline
        & Not Negative & Negative \\
\hline
True Not Negative & 2492        & 0     \\
True Negative     & 0         & 506      \\
\hline
\end{tabular}
\end{table*}

Our team also trained bilingual embeddings using a library of mutilingual unsupervised word embeddings~\cite{lample2017unsupervised}~\footnote{https://github.com/facebookresearch/MUSE} to compare to our code-switched embeddings, with the only difference in the models and training procedure being the embeddings themselves.
The results are presented in table~\ref{compare_results}.
Results are comparable; it is noteworthy that our embeddings only contain 255,062 vocabulary words, compared to the 4,027,169 words in the bilingual embeddings.
Both models had difficulties distinguishing the neutral class from the positive class, however, the model trained with code-switched embeddings was better at classifying positive tweets. 

\begin{table}[h]\caption{Code-switched and Bilingual Embedding Comparison with dimension 100}
\label{compare_results}
\center
\begin{tabular}{|l||ll|ll|ll|l|}
\hline
          & Precision && Recall && f-1 score && support \\ \hline
          & CS      &  Bilingual    & CS    & Bilingual     & CS    & Bilingual     &    \\\hline
Positive  & 0.62    &   0.62       & \textbf{0.70}      &   0.68     & \textbf{0.66}          & 0.65          & 1499    \\ \hline
Neutral   & \textbf{0.44} &   0.43       & 0.36      &   0.36     & 0.39          & 0.39          & 993     \\ \hline
Negative  & 1.00 &   1.00       & 1.00      &  1.00      & 1.00          & 1.00          & 506     \\ \hline
macro avg & \textbf{0.69} &   0.68       & 0.68      &   0.68         & 0.68          & 0.68              & 2998    \\ \hline
\end{tabular}
\end{table}

\section{Conclusion}
\label{conclusion}
In previous work, multilingual word embeddings have been used to classify code-switched text, however, these embeddings were not trained on code-switched data. In this work, we presented multilingual embeddings trained on code-switched social media text.~\footnote{https://github.com/francesita/Code-switch-Embeddings-for-Sentiment-Analysis} The results show that code-switched embeddings are able to capture the meanings of words when code-mixing occurs despite having a vocabulary much smaller in size. They also show that for the application of sentiment analysis, these embeddings are successfully employed to train a sentiment classifier, see table~\ref{task_results}. In future work, we will use Google's Bidirectional Encoder Representations from Transformer (BERT) to obtain contextualised word embeddings from code-switched data. To our knowledge, although BERT has multilingual models, there are no code-switched models. We will also continue to collect data for code-switched word embeddings in Spanglish and begin collection for Hinglish.

\bibliographystyle{coling}
\bibliography{semevalPaper}

\end{document}